\documentclass[11pt]{article}

\usepackage[final]{acl}

\usepackage{times}
\usepackage{latexsym}

\usepackage[T1]{fontenc}

\usepackage[utf8]{inputenc}

\usepackage{microtype}

\usepackage{inconsolata}

\usepackage{graphicx}

\newcommand{\KL}{D_{\mathrm{KL}}}
\usepackage{comment}
\usepackage{makecell}

\usepackage{amsmath}
\usepackage{amssymb}
\usepackage{mathtools}
\usepackage{amsthm}
\usepackage{microtype}
\usepackage{graphicx}
\usepackage{subcaption}
\usepackage{booktabs} 
\usepackage{pgfplotstable}
\usepackage{pgfplots}
\usepackage{threeparttable}
\pgfplotsset{compat=1.16} 
\usepgfplotslibrary{groupplots}
\usetikzlibrary{backgrounds, positioning}
\usepackage{hyperref}
\usepackage[section]{placeins}
\usepackage{float}
\usepackage{multirow}
\usepackage{mathrsfs} 
\usepackage{enumitem}

\usepackage{tikz}
\usepgfplotslibrary{groupplots}
\pgfplotsset{compat=1.18}

\usepackage{textcomp}  
\usepackage{scalerel}  
\def\uhel{\textsuperscript{\hspace{-0.5em}\scalerel*{\includegraphics{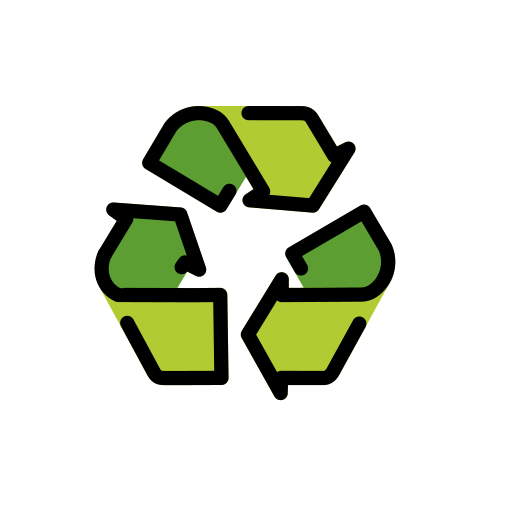}}{\textrm{\LARGE\textbigcircle}}}\hspace{-0.4em}}
\def\lisn{\textsuperscript{\hspace{-0.35em}\scalerel*{\includegraphics{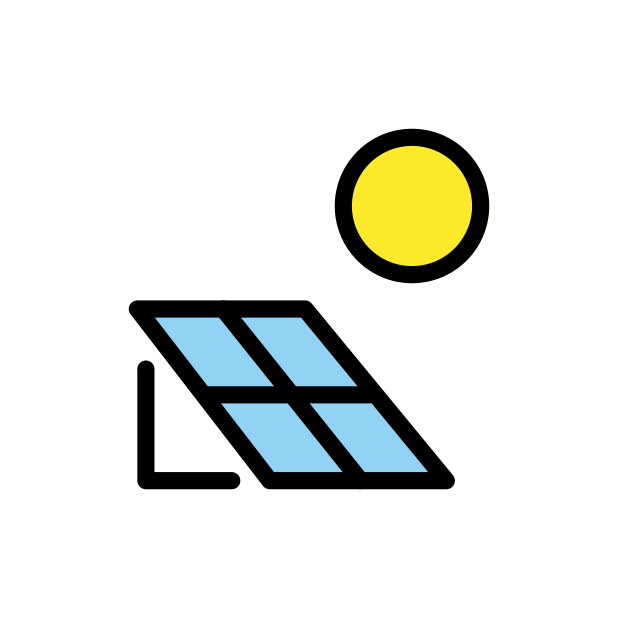}}{\textrm{\LARGE\textbigcircle}}}\hspace{-0.4em}}
\def\ensiie{\textsuperscript{\hspace{-0.5em}\scalerel*{\includegraphics{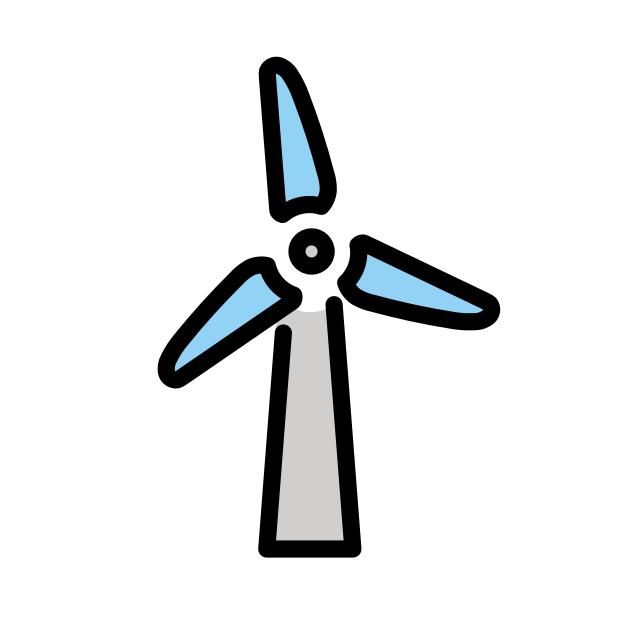}}{\textrm{\LARGE\textbigcircle}}}\hspace{-0.4em}}

\title{Is Knowledge Distillation Actually Greener? \\ A Case Study in Machine Translation}

\author{
Joseph Attieh\uhel \quad Timothee Mickus\uhel \quad Anne-Laure Ligozat\lisn\textsuperscript{\hspace{0.1em}}\ensiie \\ \textbf{Aurélie Névéol}\lisn \quad \textbf{Jörg Tiedemann}\uhel
\\[0.2em]
\uhel{}University of Helsinki \quad \lisn{}Université
Paris-Saclay, CNRS, LISN  \quad \ensiie{}ensIIE\\ \texttt{first.last@\{helsinki.fi,lisn.fr\}}
\\}

\begin{document}
\maketitle

\begin{abstract}
Knowledge distillation (KD) is a technique
to compress a
larger \textit{teacher} system into a smaller \textit{student}. In machine translation, {KD is commonly evaluated through translation quality and inference efficiency, without jointly accounting for the environmental costs of producing and deploying the distilled system.} 
We evaluate representative KD methods both on bespoke MT models and LLMs, by considering both translation quality and computational cost, 
using the Machine Learning Life Cycle Assessment tool, 
which accounts for
costs throughout the KD model life cycle.  
{Our key finding is that the deployment volume required to amortize KD is serving-dependent and can shift by several orders of magnitude under batching.}
We include actionable guidance for selecting, developing, and evaluating  KD methods under quality and compute-induced constraints.

\end{abstract}

\section{Introduction}
Knowledge distillation (KD) is a standard family of algorithms that 
transfers knowledge from a larger teacher model to a smaller student model~\cite{hinton2015distilling}.  Because students are smaller and cheaper to run, KD is presented as a way to reduce inference cost and enable deployment in resource-constrained settings~\cite{luccioni2024power}. 
However, KD introduces costs that are often ignored. Compressing the model can reduce translation quality relative to the teacher~\cite{gumma-etal-2023-empirical}, and adds compute costs beyond standard training~\cite{zhang-etal-2018-analyzing}:   sequence-level KD requires inference from the teacher over the full dataset, whereas logit-level approaches require repeated teacher forward passes or precomputed teacher distributions. {Although prior work has quantified selected computational or environmental costs of KD, such analyses typically cover only parts of the distillation and deployment life cycle and  often emphasize task quality or inference efficiency.}  
As a result, existing evaluations provide an incomplete picture of the process, 
making it difficult to decide whether KD should be applied or which KD method should be used. 
\begin{figure}[t]
    \centering
    \includegraphics[width=0.98\columnwidth]{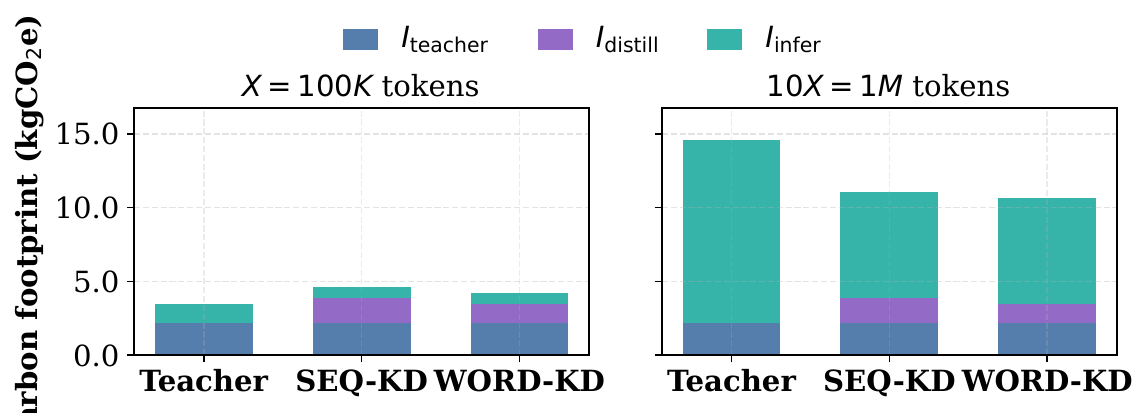}
    \caption{Motivating footprint decomposition for 16M student on EN$\rightarrow$IS. At low served volume $X$, teacher training and distillation costs dominate KD footprints; at larger volume, inference costs dominate and cheap  student inference can offset distillation overhead. For scale, $X$ roughly corresponds to the number of words in the novel \emph{Harry Potter and the Prisoner of Azkaban}.}
    \label{fig:motivation_decomposition}
\end{figure}
Figure~\ref{fig:motivation_decomposition} shows the central trade-off: KD can increase emissions at low deployment volume 
as teacher training and distillation are fixed production costs, whereas compact students can become favorable once inference dominates. 
This frames KD selection as an amortization problem: KD is not automatically greener because the student is smaller; KD is only environmentally beneficial if the student 
serves enough volume to offset its production footprint.

%
We study this question in machine translation (MT). 
MT offers the benefits of standardized datasets, established evaluation metrics, and reproducible training pipelines, allowing us to isolate the life cycle impact of KD. 
We adopt the Machine Learning Life Cycle Assessment framework (MLCA, \citealp{morand2024}) to evaluate the environmental impact of performing KD 
in terms of carbon footprint. 
Our contributions are as follows:
\begin{enumerate}[nosep]
    \item  We propose a life cycle aware evaluation protocol for KD in MT, decomposing environmental impact into teacher training, distillation, and inference, while accounting for operational and amortized embodied emissions.
    \item We benchmark representative KD methods by evaluating \emph{both} translation quality and carbon impact, emphasizing stark differences in carbon impact for equal-performance methods. 
    \item We evaluate across student capacity, language pair, corpus size, and teacher-cost accounting. 
    We experiment with both bespoke encoder--decoder MT models and decoder-only LLMs. 
    \item We quantify deployment-volume break-even points and show that they are serving-dependent and can shift by several orders of magnitude under batching.

\end{enumerate}


\section{Life Cycle Accounting for Distillation}
\label{sec:accounting}
This section introduces the distillation regimes and carbon-accounting framework used throughout the paper. We first distinguish the two dominant KD families considered in our experiments and then describe how their training, preprocessing, and inference costs are incorporated into our life cycle assessment.
\subsection{Distillation Methods and Life Cycle Costs}
\label{sec:KD}
We compare the two dominant families of KD for MT: sequence-level KD and word-level KD. Sequence-level KD, commonly denoted as Seq-KD~\cite{kim-rush-2016-sequence} first decodes synthetic targets from the teacher and then trains the student on them, so its additional cost comes mainly from teacher-side generation (often with beam search). Word-level KD, referred to as Word-KD~\cite{kim-rush-2016-sequence}, instead trains the student to match the teacher's token-level predictive distribution on the training targets. In its online form, this requires repeated teacher forward passes during student training and access to teacher logits. Offline variants also exist: they precompute and store teacher distributions, often as top-$K$ logits or probabilities, before student training. This shifts part of the Word-KD cost from training to teacher-side preprocessing and storage, but it does not remove the need to run the teacher over the distillation corpus. 

These two regimes therefore create different life cycle profiles: Seq-KD pays an upfront synthetic-generation cost, whereas Word-KD pays for teacher scoring either during training or in a preprocessing stage.  See Appendix~\ref{app:kd_objectives} for objectives and notations.

\subsection{How to Assess Costs using MLCA}
\label{sec:mlca}
\paragraph{Methodology.}
To estimate the environmental impact of ML solutions throughout their entire life cycle, we adopt the Machine Learning Life Cycle Assessment framing (MLCA, \citealp{morand2024}), which adapts the Life Cycle Assessment methodology (LCA) for ML systems. MLCA aims at being as comprehensive as possible 
by accounting for data center overheads and embodied emissions from hardware manufacturing. Unlike FLOPs, MLCA estimates the environmental impact of executing computation by accounting for runtime, hardware utilization, data-center overhead, electricity mix, and embodied hardware impacts; similar FLOP counts do not imply similar carbon footprints~\citep{fernandez-etal-2025-energy}.

MLCA decomposes the assessment into phases through which a product or service incurs environmental impacts, from a software and a hardware point of view. From a \textbf{software point of view}, the phases correspond to the different tasks involved in the ML service, such as data acquisition, storage, training, and inference. From a \textbf{hardware point of view}, it includes mainly   (i) production (raw material extraction, manufacturing and transport of hardware), (ii) use (most of the electricity and energy consumption of the equipment 
while the hardware runs the tasks), and (iii) end-of-life (the processes of reuse, recycling, or disposal of the hardware)~\cite{ligozat2022unraveling}.
 
 Each phase of the life cycle has impacts on different environmental indicators, such as greenhouse gas emissions, water footprint, or abiotic resource depletion.  Since our study focuses on reducing 
 energy consumption, 
 the main gain is expected to be in terms of carbon footprint~\cite{ligozat2022unraveling}. Therefore, we assess greenhouse-gas emissions in CO$_2$-equivalents (CO$_2$e), a standard unit that converts the warming effect of different greenhouse gases into the equivalent amount of CO$_2$. 

 \paragraph{Functional unit and System Boundary. }
 MLCA requires defining a functional unit (FU) and a system boundary~\cite{ISO14040,ISO14044}. The FU specifies the question the MLCA aims to answer, while the system boundary defines the scope of impacts assessed. Our FU is defined as \emph{``What are the impacts of producing an MT system that serves $X$ translation requests (or decoded target tokens) over one year at a specified quality level?"}

\begin{figure}[t]
  \centering
  \includegraphics[width=0.9\linewidth]{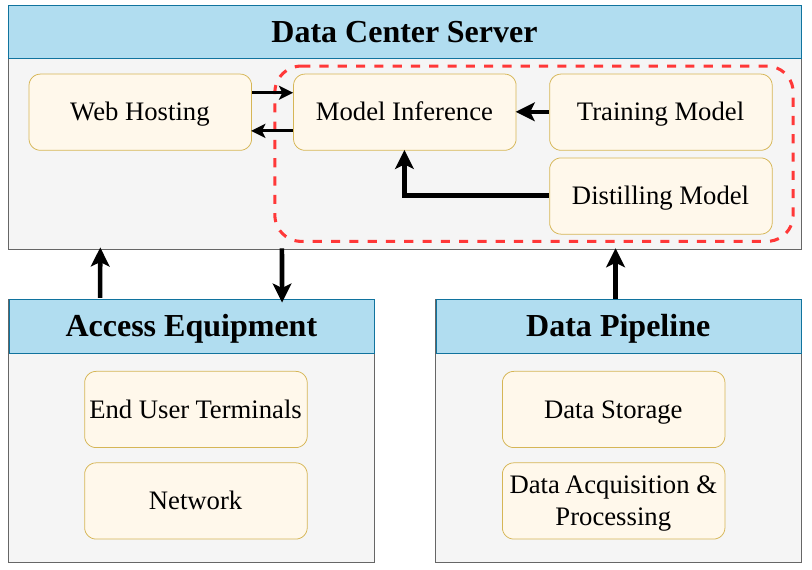}
  
  \caption{System boundary of the MLCA highlighted by the red dashed line.}
  \label{fig:boundary}
\end{figure}

From a \textbf{software point of view}, the boundary used in this study is restricted to the data-center server side of the KD pipeline as shown in Figure~\ref{fig:boundary}. We define the compute phases $\phi \in \{\textsc{train},\textsc{distill},\textsc{infer}\}$. End-user devices, network transfer, and data acquisition/storage are excluded, as the study adopts a compute-centric system boundary; these components are outside the scope and assumed invariant across the systems compared.  
From a \textbf{hardware point of view}, we cover \textit{embodied (production)}  and \textit{operational (use-phase) impacts} as recommended by MLCA; the end-of-life phase is difficult to estimate due to limited data on the end-of-life treatment of most equipment~\cite{ficher_marion}. Appendix~\ref{app:carbon_details} reports the full accounting assumptions.

\paragraph{Phase-level carbon accounting.}
For each 
phase $\phi 
$ and device $e$, we estimate 
footprint as the sum of operational and amortized embodied emissions~\cite{luccioni2024power,berthelot2024}:

\begin{align*}
I^{(\phi)}_e  &=  \underbrace{ \mathrm{PUE}\;\times\;E^{(\phi)}_e\;\times\;\mathrm{EGM}_g }_{\text{Operational Emissions}} \\
   &\qquad + \underbrace{
F^{(\mathrm{prod})}_e\;
\frac{t^{(\phi)}_e}{t_{\mathrm{life},e}\times \mathrm{AUR}_e} }_{\text{Embodied Emissions}}
\end{align*}

Here, $E^{(\phi)}_e$ is the electricity consumed by phase $\phi$, $\mathrm{PUE}$ accounts for data-center overhead, $\mathrm{EGM}_g$ is the grid emission factor, $F^{(\mathrm{prod})}_e$ is the manufacturing footprint of device $e$, $t^{(\phi)}_e$ is the phase runtime, $t_{\mathrm{life},e}$ is hardware lifetime, and $\mathrm{AUR}_e$ is the active utilization rate.  The total footprint of a system is obtained by summing 
over phases and devices. 
See 
Appendix~\ref{app:carbon_details} for details.

\paragraph{Teacher-cost allocation.}
A central modeling choice in life cycle accounting for KD is what fraction of  the teacher production footprint $\lambda_T \in [0,1]$ should be assigned to a distilled student. 
In the main MT experiments, we use a conservative teacher-included allocation with $\lambda_T = 1$: the teacher training footprint is assigned to the distilled system. This reflects a setting where the teacher is solely trained for the purpose of producing a deployable student.\footnote{If the same teacher is reused to produce $K$ students, its production footprint could instead be amortized with $\lambda_T = 1/K$, lowering the fixed cost of each distilled system.} In the LLM case study, we use sunk teacher-cost allocation, i.e., $\lambda_T = 0$, since the exact pretraining footprint of the LLM teacher is unavailable and cannot be reconstructed reliably.
\begin{figure*}[t]
    \centering
    \includegraphics[width=\textwidth]{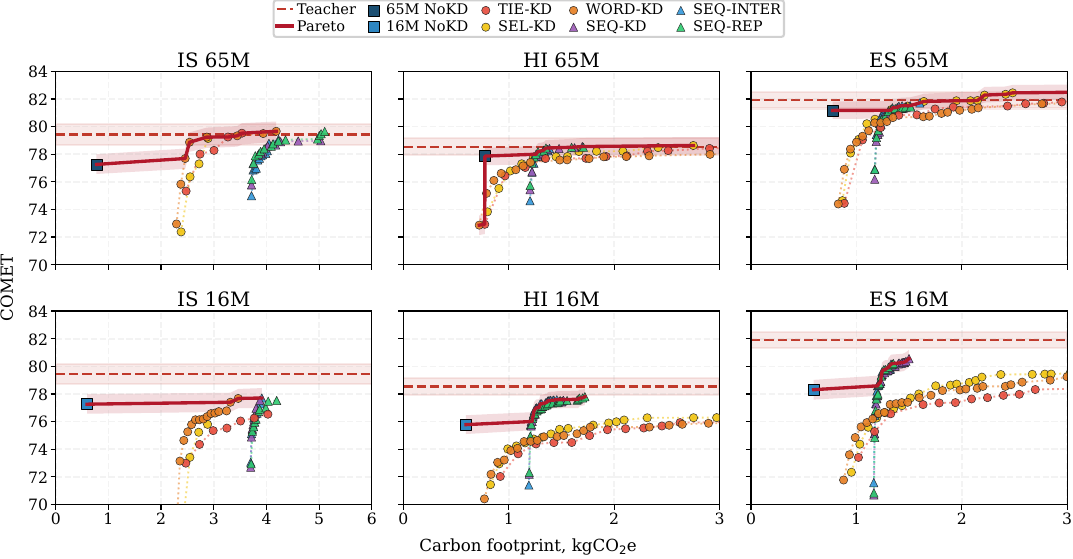}
    \caption{ Global Pareto frontiers across student sizes and language pairs, in terms of COMET and production footprint of the models (i.e., footprint of training the teacher and distilling). Shaded regions show COMET confidence intervals (paired bootstrap, $N=1000$, 95\% CI); dashed lines denote teacher COMET.}
    \label{fig:pareto-frontiers}
\end{figure*}

\paragraph{Break-even deployment volume.}
For a deployed system $m$, we write total footprint at served token volume $X$ as
$I_m(X) = I^{m}_{\mathrm{prod}} + X \cdot c^{m}_{\mathrm{infer}}$,
where $I^{m}_{\mathrm{prod}}$ is the fixed production footprint, including the allocated teacher cost and any distillation overhead, and $c^{m}_{\mathrm{infer}}$ is the per-token inference footprint. For a distilled student $S$ compared with a teacher $T$, the break-even volume can be trivially computed by solving for $X$. Below that volume, the fixed cost of distillation is not amortized. 
{Since KD and No-KD deploy the same student architecture, they have the same per-token inference footprints and therefore no deployment-volume break-even. Their comparison is instead quality-constrained: KD is justified only if its additional production footprint yields the required quality gain.}

Although this amortization relation is simple, its practical value depends on quantities that are rarely reported in efficiency-oriented KD work: the fixed production footprint $I_{\mathrm{prod}}$, the per-token inference slope $c_{\mathrm{infer}}$, and the resulting break-even volume $X^\star$ across methods. Our contribution is therefore not the identity itself, but its empirical instantiation in a controlled MT setting. 

This amortization analysis does not require retraining models for every deployment scenario. Once training, distillation, and inference phases are profiled on a given hardware stack, their energy and emissions can be estimated from a small number of measurements and simple scaling with served volume.\footnote{This is consistent with large-scale empirical studies such as \citet{luccioni2023}.} 
The computational cost of running the amortization analysis itself is negligible compared with training and distillation, and does not materially affect the conclusions.  Consequently, this profiling framework can be seamlessly integrated into standard model evaluation pipelines. 

\section{Experimental Setup}
\label{sec:methodology}

We benchmark MT distillation methods jointly in terms of translation quality and carbon impact.  To keep comparisons focused and reproducible, we fix the data sources, tokenization, model family, optimization setup, and evaluation protocol across methods, and vary only the KD method, student capacity{, and language pair}. 

See Appendices~\ref{app:data_artifacts} and ~\ref{app:training_details} for data documentation, model configurations, software details, and computational budget.

\paragraph{Data and Evaluation}
\label{sec:setup:data}
We use the parallel corpora released for the \textsc{WMT 2024} General MT constrained track~\cite{bojar-EtAl:2014:W14-33}.\footnote{\url{https://www2.statmt.org/wmt24/mtdata/}} We evaluate three translation directions, from English (EN) to Icelandic (IS), Hindi (HI)  and Spanish (ES). These directions differ in corpus size: EN$\rightarrow$IS uses the released 23M sentence pairs, while EN$\rightarrow$HI and EN$\rightarrow$ES are uniformly sampled to 13M and 6M sentence pairs, respectively. This allows us to study how corpus size affects the production footprint of distillation and the resulting footprint--quality trade-offs.
We evaluate {translation quality} on FLORES+ devtest~\cite{gordeev-etal-2024-flores} and report COMET~\cite{rei-etal-2020-comet}.\footnote{We use \texttt{Unbabel/wmt22-comet-da}.}

We distill on the same data used for training the teacher.
While distillation does not require the teacher and student to be trained on identical corpora~\cite{aji2021fullysyntheticdataimproves}, this allows us to isolate the effect of the KD method and factor out the impact of dataset choice.

\paragraph{Models, Baselines, and KD Methods}
\label{sec:setup:models}
We adopt the Transformer (TF) architecture~\cite{vaswani2027} for both teachers and students. We use a TF-Big teacher and two student architectures: TF-Base (moderate compression) and TF-Tiny (aggressive compression). Architectural details and parameter counts are reported in Table~\ref{tab:arch} in Appendix~\ref{app:training_details}.
{For each language pair and student size, we compare three deployment scenarios}:
(i) \textbf{No-KD baseline}, where a compact model (i.e., the same size as the student) is trained directly on parallel data and used for all inference, (ii) \textbf{Teacher baseline}, where a large model is trained and used for all inference, and
(iii) \textbf{KD pipeline}, where a large teacher is trained, a smaller student is distilled, and inference is served by the student.  %
{ We evaluate representative KD methods from the two dominant KD regimes described in detail in Appendix~\ref{app:kd_objectives}. 
}
In this study, we evaluate variants of \textbf{Word-KD}~\cite{kim-rush-2016-sequence}: Sel-KD~\cite{wang-etal-2021-selective}, and TIE-KD~\cite{zhang-etal-2023-towards-understanding}; and \textbf{Seq-KD}~\cite{kim-rush-2016-sequence}: Seq-Inter~\cite{kim-rush-2016-sequence}, and Seq-Rep~\cite{zhang-etal-2018-analyzing}; all described in Section~\ref{sec:related} and Appendix~\ref{app:kd_objectives}.  

\paragraph{Carbon Footprint Measurement}
\label{sec:setup:carbon}
All experiments are run on a single NVIDIA V100 GPU to keep comparisons controlled and sidestep distributed training overhead.\footnote{We tested 4-GPU distillation; in our setting, the additional device footprint outweighed runtime gains, making the configuration dominated by single-GPU distillation (Appendix~\ref{app:distributed}).}  For each phase, we log wall-clock time, utilization statistics, and energy use, enabling comparable energy and carbon estimation across KD methods.  We profile teacher training, distillation, and inference separately so that the footprint of each phase can be attributed to the corresponding system component. 
See Appendix~\ref{app:carbon_details} for parameters selected and sensitivity analysis. 

\section{Results}
\label{sec:results}

Figure~\ref{fig:motivation_decomposition} shows the basic amortization pattern; here, we focus on two questions: which KD methods lie on the footprint--quality frontier, and at what served volume their production footprint is amortized. Appendix~\ref{app:full_results} reports complementary results to this section, including footprint decompositions and break-even values.

\subsection{What KD method should we pick?} 
\label{sec:rq3_method_choice}

We first assess which KD methods are worth producing before considering deployment volume. Method selection is framed as a trade-off between translation quality and the upfront production footprint, which includes teacher training and KD overhead but excludes inference.



Figure~\ref{fig:pareto-frontiers} shows footprint--quality Pareto frontiers. Each point corresponds to a checkpoint, with production footprint plotted on the $x$-axis and COMET on the $y$-axis. The \emph{global Pareto frontier} highlights configurations for which no other point achieves both higher COMET and lower footprint, helping avoid suboptimal quality--impact trade-offs.


The No-KD baseline has the lowest production footprint in general because it avoids teacher training allocation and distillation overhead. 
{Therefore, KD is worthwhile over No-KD only when the quality gain justifies its additional production footprint; marginal COMET improvements may therefore be unattractive when the No-KD model already approaches the required quality level.}
{
Figure~\ref{fig:pareto-frontiers} shows that KD method choice depends on student capacity, language pair, and dataset size. The latter directly shifts the production-footprint axis: EN$\rightarrow$IS  uses the largest corpus, so teacher decoding is most expensive, while the smaller datasets reduce the upfront costs of distillation. This strongly impacts Seq-KD variants, whose target-generation cost scales with the number of source tokens decoded by the teacher.
For the 65M student, word-level methods generally provide the best trade-offs, often reaching 
teacher-level quality. In this moderate-compression regime, token-level supervision is often sufficient to recover quality while avoiding the fixed decoding cost of Seq-KD.  
For the 16M student, capacity becomes a stronger bottleneck, and Seq-KD methods can become more competitive when synthetic targets simplify the learning problem. This effect is clearer on the smaller datasets, where the target-generation footprint is lower. However, EN$\rightarrow$IS shows the limits of this strategy: here, 
the 16M student gains little over No-KD and remains below the teacher; additional distillation yields diminishing returns.}

Overall, no KD method is universally dominant. Word-KD  variants are generally preferable for moderate compression while Seq-KD variants become attractive for smaller students when their quality justifies the synthetic generation footprint. Dataset size not only rescales the carbon footprint; it can also change which KD family is competitive.  



\subsection{When do the inference costs justify KD?}
\begin{figure}[t]
    \centering
    \includegraphics[width=\linewidth]{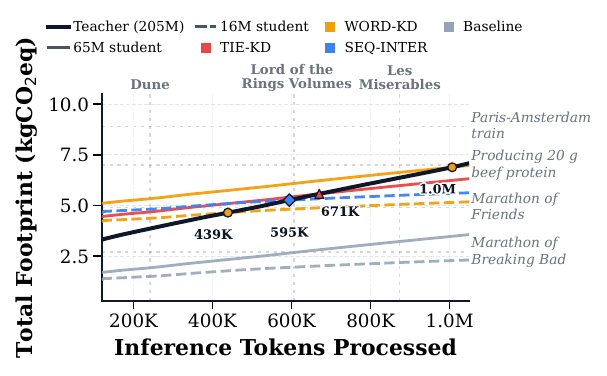}
    \caption[Amortization of distillation cost vs. inference volume.]{
    Amortization of distillation cost vs.\ inference volume {under unbatched sentence-level inference} for $EN\rightarrow IS$.
    Markers denote break-even points where students become less costly.
    Dotted horizontal guides contextualize the footprint. 
    Dotted vertical guides contextualize token volume.\protect\footnotemark
    }
    \label{fig:payback}
\end{figure}

The decomposition in Figure~\ref{fig:motivation_decomposition} suggests that KD becomes environmentally viable only above a workload threshold.
To determine this threshold, we turn our life cycle accounting into an amortization problem,  
 where we model the inference impact of a deployed model to be approximately linear in the number of translated tokens.

\paragraph{Sentence-level Inference}

We first report inference costs under a simplified serving setting, where the model translates one sentence at a time. Figure~\ref{fig:payback} shows that KD becomes environmentally preferable only after a specific usage threshold. 
KD students have a higher intercept than the teacher because of distillation overhead, but a lower slope because student inference is cheaper. Below the break-even point, fixed production costs dominate; above it, inference savings can compensate for the upfront cost. 
The 16M student breaks even earlier because of its lower inference slope,  but this comes at a stronger quality ceiling as shown in the previous section. Seq-KD variants usually require larger served volumes because teacher decoding increases $I_{\mathrm{prod}}$ without changing inference cost. 
In sum, KD is only beneficial if the student satisfies the quality requirement and is deployed at sufficient volume.

\footnotetext{
All contextual guides are illustrative and depend on device, electricity mix, preprocessing, and tokenizer.
}
{
\paragraph{Batch Inference}
\label{app:inference_scaling}
 MT systems commonly batch requests to improve hardware utilization. To quantify the effect on amortization, we measure inference footprint over fixed workloads of 32K--1M target tokens across multiple batch sizes, repeating each setting three times, and fit the resulting footprint--volume relationship for each model and serving configuration. We then combine these inference curves with the fixed production footprints to compute break-even volume against the teacher.}

\begin{table}[h]
\centering
\resizebox{\columnwidth}{!}{
\begin{tabular}{lcccc}
\toprule
& \multicolumn{2}{c}{16M} & \multicolumn{2}{c}{65M} \\
\cmidrule(lr){2-3} \cmidrule(lr){4-5}
Batch size
& SEQ-INTER & WORD-KD
& TIE-KD & WORD-KD \\
\midrule
256   & 386.97M & 285.07M & 733.37M & 1.09B \\
1024  & 973.63M & 717.23M & 2.34B   & 3.49B \\
4096  & 1.28B   & 942.02M & 1.38B   & 2.06B \\
16000 & 1.60B   & 1.18B   & 2.51B   & 3.75B \\
\bottomrule
\end{tabular}}
\caption{Break-even inference volume against the teacher under different
batch sizes for selected Pareto-optimal configurations ($EN \rightarrow IS$).}
\label{tab:payback_new}
\end{table}

{
Table~\ref{tab:payback_new} shows that batching shifts break-even from the token range observed under sentence-level inference to hundreds of millions or billions of tokens, as it improves both teacher and student inference, reducing the student's relative per-token advantage. The trend is not perfectly monotonic across batch sizes as measured utilization and runtime vary across configurations. Break-even should therefore be estimated under the target serving configuration rather than treated as an intrinsic property of a distilled model.
}


\subsection{What about LLM-based MT?}
\begin{table*}[t]
\centering
{\scriptsize
\begin{tabular}{llcclrrrrc}
\toprule
\multirow{2}{*}{Teacher} 
& \multirow{2}{*}{Pair}
& \multicolumn{5}{c}{COMET score}
& \multicolumn{2}{c}{$I^{distill}$(gCO$_2$e)} 
& \multirow{2}{*}{\makecell[c]{Break-Even\\Tokens (M)}} \\
\cmidrule(lr){3-7} \cmidrule(lr){8-9} 
& & Teacher & Base & Method & Student & $\Delta$ Base 
& KD train & Distilled data & \\
\midrule
 
\multirow{4}{*}{Qwen3-8B}
& \multirow{2}{*}{EN$\rightarrow$ES}
& \multirow{2}{*}{86.28}
& \multirow{2}{*}{76.06}
& Seq-KD  & 77.88 & +1.82 & 18.76 & \multirow{2}{*}{1047.58} & 12.34 \\
& & & & Word-KD & 80.76 & +4.70 & 91.97 & &  13.19 \\
\cmidrule(lr){2-9}

& \multirow{2}{*}{EN$\rightarrow$IS}
& \multirow{2}{*}{51.77}
& \multirow{2}{*}{28.30}
& Seq-KD  & 48.18 & +19.88 & 48.96 & \multirow{2}{*}{854.22}  & 10.45\\
& & & & Word-KD & 47.99 & +19.69 & 79.53 &  & 10.81\\

\midrule

\multirow{4}{*}{Qwen3-4B}
& \multirow{2}{*}{EN$\rightarrow$ES}
& \multirow{2}{*}{85.37}
& \multirow{2}{*}{76.06}
& Seq-KD  & 78.02 & +1.96 & 17.31 & \multirow{2}{*}{983.10} & 11.05\\
& & & & Word-KD & 80.06 & +4.00 & 97.48 & & 11.94 \\
\cmidrule(lr){2-9}

& \multirow{2}{*}{EN$\rightarrow$IS}
& \multirow{2}{*}{42.41}
& \multirow{2}{*}{28.30}
& Seq-KD  & 38.99 & +10.69 & 25.73 & \multirow{2}{*}{756.72} & 8.65\\
& & & & Word-KD & 46.36 & +18.06 & 80.92 &  & 9.26 \\

\bottomrule
\end{tabular}}
\caption{Qwen3 distillation results under sunk teacher-cost allocation. 
KD train is the carbon footprint of student distillation. 
Distilled data tracks target generation for Seq-KD and offline teacher-distribution compute for Word-KD.}
\label{tab:qwen_llm_results}
\end{table*}

Finally, we test whether the same life cycle logic holds in a decoder-only LLM setting. Unlike the encoder--decoder experiments, we use sunk teacher-cost accounting ($\lambda_T=0$), since the pretraining footprint of LLM teachers is generally unavailable. This scoped case study tests whether the same life cycle logic holds under more favorable assumptions and is not intended as a complete benchmark of LLM distillation for MT. 

We distill Qwen3-8B~\cite{yang2025qwen3technicalreport}\footnote{ \texttt{https://huggingface.co/collections/Qwen/qwen3}} and Qwen3-4B teachers into a Qwen3-0.6B student on EN$\rightarrow$IS and EN$\rightarrow$ES using 100K sampled sentence pairs.\footnote{Word-/-logit-based KD 
requires teacher and student models to share vocabulary, which is the case in the Qwen3 family.} 
For Seq-KD, the teacher generates synthetic translations. Unlike the previous experiments,  Word-KD is offline; therefore, teacher token distributions are precomputed before student training and then reused during distillation. Full details are reported in Appendix~\ref{app:llm_details}. This provides a complementary evaluation setting with a different model architecture and distillation implementation from the main experiments.

The results support the same conclusion as the main MT experiments. Distillation improves over the Qwen3-0.6B baseline in all settings, especially for EN$\rightarrow$IS, where the baseline is weak.
{Notably, the Qwen3-4B Word-KD student outperforms its teacher on EN$\rightarrow$IS. This may reflect task-specific adaptation of the student and the regularizing effect of KD rather than greater general capability~\citep{xu-etal-2025-self-distillation,gordon2019explainingsequencelevelknowledgedistillation}.}

However, even under sunk teacher-cost accounting, KD still incurs measurable production costs from teacher-side preprocessing and student training. Offline Word-KD is a higher-supervision but higher-cost regime: it gives the student richer token-level teacher information than Seq-KD, but it also requires offline teacher-distribution extraction and leads to a higher student-training footprint. In most settings, this extra cost buys larger COMET gains; however, for Qwen3-8B on EN$\rightarrow$IS, offline Word-KD costs more while giving similar COMET to Seq-KD. Thus, offline Word-KD is not uniformly preferable.  More broadly, lower inference cost alone is not sufficient: LLM distillation should be evaluated jointly through quality, production footprint, and expected deployment volume.

\section{Discussion}
\label{sec:discussion}

A key takeaway {from our machine translation experiments} is that the perceived efficiency of KD depends on what is used to evaluate the system. If we only consider inference costs, distillation looks uniformly beneficial because the cost of handling a request with a student model is lower. However, under MLCA accounting, KD introduces a fixed, upfront footprint (encompassing both teacher training and distillation overhead) that dominates total emissions at low served volume $X$ (Appendix~\ref{sec:rq1_cost}). {The resulting quality–footprint trade-offs and break-even volumes are specific to the MT settings studied here. The life cycle accounting framework itself is not MT-specific and can be applied to other KD settings, although these quantities must be re-estimated for each task and deployment regime.}

Therefore, 
whether to use KD
fundamentally depends on the deployment: 
cost savings are not inherent to KD, but depend on the downstream workload.
Studies that report only translation quality (or only inference efficiency) can lead to systematically different conclusions under life cycle-aware evaluation.  Benchmarking KD under an explicit functional unit tied to served volume makes the trade-off operational: 
{for MT,} KD is green
only when deployed at 
scale, under a quality constraint, and with a method whose
distillation overhead does not erase its inference savings.

Across settings, we observe a consistent shift. For small $X$, total emissions are driven by one-time costs,
so pipelines without distillation overhead (Teacher-only, No-KD) have lower carbon footprint. As $X$ grows, the inference term dominates, and smaller students gain an advantage through a lower per-token slope; KD becomes favorable once inference savings amortize the initial overhead. 
Practically, this means that for one-off translation (single document / single novel-scale usage),
KD is rarely justified from an emissions perspective, whereas for sustained usage (repeated service or high-throughput pipelines), KD can reduce life cycle emissions while meeting a quality constraint. 

{
In practice, whether KD becomes environmentally preferable at a realistic deployment volume depends strongly on the serving configuration. Under sentence-level inference, break-even can occur at relatively modest volumes, whereas batching can shift the threshold to hundreds of millions or billions of target tokens. Break-even volume should therefore be estimated under the expected production serving regime rather than inferred from model size or unbatched measurements alone.}
Student size controls both break-even volume and attainable quality. Extreme compression lowers inference costs and shifts break-even to smaller $X$, but our 16M results suggest the presence of a capacity bottleneck. The 65M student better preserves teacher-level quality while retaining substantially lower inference emissions than the teacher, yielding the strongest quality–amortization balance. 
This highlights the trade-off between the translation quality and the number of parameters of the models. 

The LLM case study supports the same conclusion under a more favorable sunk teacher-cost allocation. Excluding teacher pretraining lowers the fixed cost assigned to KD, but does not remove teacher-side preprocessing or student-training costs. In this setting, offline Word-KD provides richer token-level supervision and often improves COMET more than Seq-KD, but it also incurs a higher production footprint. Thus, it is worthwhile only when the quality gain justifies the additional cost.

Our analysis assumes full access to the teacher to either decode synthetic targets (Seq-KD) or query token distributions during training (Word-KD). This assumption is not valid for closed-source or hosted models, where logits are unavailable and teacher queries are rate-limited. In such settings, Seq-KD variants are the only viable distillation methods. This raises a broader question about environmental responsibility under access constraints. When Word-level KD methods are infeasible, higher upfront emissions of Seq-KD become an unavoidable cost of relying on closed-source teachers.  From a life cycle perspective, this shifts part of the environmental burden from algorithmic choice to access policy. Restricting access to the model may implicitly force practitioners toward higher-emission distillation pipelines, even when lower-footprint alternatives would be preferable under open-teacher assumptions. Quantifying and attributing this constrained-choice penalty is an important future direction. 

\paragraph{Actionable guidance for {MT} practitioners.}
Our results suggest a simple decision procedure:
    
    \noindent \textbf{1. 
    Decide
    with a functional unit.} Specify the intended {translation token} volume $X$; if low, use the teacher. 
    KD  
    lowers impact only if its upfront footprint is amortized by inference servings at the target $X$. 
    
    \noindent \textbf{2. Start with a baseline}.
    Establish a No-KD {MT} baseline to determine the minimum footprint system and quantify the quality gap KD must justify.
    
    \noindent \textbf{3. Prefer low-overhead KD to reduce upfront emissions.} When KD is required to meet a {translation} quality target 
    with the lowest production footprint, avoid sequence-level methods unless 
    they meet 
    foreseeable
    quality gains or operational constraints. 
    
    \noindent \textbf{4. Report and select from Pareto-optimal configurations.} Treat model selection as a multi-objective problem: choose checkpoints on the global footprint--quality Pareto frontier and report uncertainty (e.g., confidence intervals), so as to prevent future work from incurring extra emissions without meaningful quality improvements.
    
    \noindent \textbf{5. 
    Report environmental impact} alongside task quality when presenting a distilled model or a new KD method. This enables a clear comparison with prior works and strengthens claims of performance when newly devised methods result in better students. 
    It also allows practitioners and users to align model choices with 
    sustainability constraints, and not rely solely on task performance metrics.


\section{Related Work}
\label{sec:related}

\paragraph{Knowledge Distillation for Machine Translation}
KD is widely used to compress neural MT systems, so as to retain a large teacher model’s translation quality while reducing inference costs~\cite{kim-rush-2016-sequence}. 
Methods mostly fall into two main regimes: sequence- and word-level distillation.

Seq-KD trains the student on teacher-generated translations rather than on gold references~\cite{kim-rush-2016-sequence}. Teacher outputs 
provide smoother targets, improving student generalization, but can also propagate errors, motivating selection and filtering strategies. 
Common variants include choosing the best teacher hypothesis via (i) sequence interpolation (Seq-Inter; \citealp{kim-rush-2016-sequence}), by generating multiple hypotheses per source sentence and selecting the candidate that best aligns with the reference typically using sentence-level BLEU or (ii) sequence replacement (Seq-Rep; \citealp{zhang-etal-2018-analyzing}) by replacing low-quality synthetic targets with reference translations.

Word-KD aligns student and teacher token distributions via KL divergence. Though effective, 
recent work has shifted toward selective distillation to mitigate noise transfer.
\citet{wang-etal-2021-selective} estimate the difficulty of a token and perform KD supervision only on this token (Sel-KD; \citealp{wang-etal-2021-selective}), while TIE-KD (\citealp{zhang-etal-2023-towards-understanding}) focuses on the teacher’s most confident predictions. 


Remarkably in MT, KD methods 
are 
mostly
only evaluated in terms of translation quality, while only accounting for the inference costs with respect to the teacher.  

{Prior work has begun to quantify aspects of the computational and environmental costs of KD, including training energy, carbon emissions, and inference efficiency.  In MT, ~\citet{zhang-etal-2023-towards-understanding} improve transparency by reporting BLEU alongside training-time overhead for their KD method, but do not quantify the environmental impact of distillation or compare life cycle costs across KD methods. Beyond MT,
\citet{niemann2025} report training and inference energy in semantic segmentation, but do not account for embodied hardware emissions. \citet{rafat2023} quantify GFLOPs, energy consumption, and carbon emissions associated with KD, but do not quantify downstream inference costs. \citet{yuan2024} compare the inference energy, runtime, and resource utilization of original and distilled models, but do not measure the costs of distillation itself. These studies therefore consider only subsets of the distillation and deployment life cycle. To our knowledge, prior work has not jointly accounted for allocated teacher production, teacher-side generation or logit extraction, student training, and inference within a common life-cycle framework that includes both operational and amortized embodied emissions, nor explicitly related these quality–footprint trade-offs to deployment volume through break-even analysis.  }
This motivates an accounting framework that captures environmental impacts across both model production and deployment.

\paragraph{Environmental Impact Accounting in ML}

This gap in understanding the costs of KD
is particularly surprising given the growing prominence of Green AI. Concerns about the environmental impact of training large machine learning models have motivated calls to report computational and carbon costs alongside performance metrics~\cite{strubell-etal-2019-energy}. Existing tools estimate energy consumption through hardware telemetry~\cite{anthony2020carbontracker,benoit_courty_2024_11171501,henderson2020systematic} or by modeling it from hardware characteristics such as TDP and runtime~\cite{lannelongue2021greenalgorithms,lacoste2019quantifyingcarbonemissionsmachine}. Most of this work emphasizes operational energy and CO$_2$e emissions during training. This narrow scope overlooks embodied impacts from hardware manufacturing and data-center overheads, which can be substantial~\cite{gupta2022chasingcarbon,luccioni2023}. For example, \citet{luccioni2023} showed that accounting for hardware production and infrastructure significantly alters BLOOM’s total footprint. 

Machine Learning Life Cycle Assessment (MLCA; \citealp{morand2024}) adapts LCA to ML by accounting for impacts across the model's entire life cycle. MLCA integrates operational energy estimates with amortized production impacts of compute hardware and reports across multiple impact categories beyond CO$_2$e. However, multiple challenges to this accounting persist, 
due to
limited data availability and uncertainty for specialized hardware and end-of-life scenarios~\cite{ficher_marion}.
\citet{berthelot2024} apply a similar LCA methodology for the digital service supporting deployed AI systems by treating the system as a multi-component service encompassing terminals, networks, hosting, model training and inference, and data management, 
and propose multi-criteria reporting 
with allocation rules that separate operational energy from embodied impacts.


\section{Conclusion}
KD is widely adopted to reduce the inference cost of MT systems, yet it is commonly evaluated through translation quality alone. We argued that this practice can be misleading under sustainability constraints, as KD introduces a non-negligible upfront footprint that must be amortized by downstream usage. To make this trade-off explicit, we employed the MLCA methodology for MT distillation, decomposing emissions into teacher training, distillation and inference, and accounting for both operational and amortized embodied emissions. 

Our results suggest that KD use hinges on 
(1) whether the target translation quality can be achieved without distilling 
and (2) the target token volume expected to be translated by the student model.   
We recommend to train and assess a No-KD baseline (lowest carbon footprint and baseline for COMET). If this baseline does not meet the quality level, then we recommend to consider applying KD by assessing the volume of inference expected from the model.  In practice, if the student model serves a small volume of tokens,  KD is not environmentally justified, and we recommend to use the teacher instead. At larger volumes, KD can reduce life cycle emissions beyond a task-dependent break-even point, turning distillation into an amortization problem. As for selecting an appropriate KD method, our results show that no single KD family dominates across all settings. Word-level methods often provide strong footprint--quality trade-offs under moderate compression, while sequence-level methods can become competitive for aggressively compressed students when synthetic targets yield larger quality gains.
Finally, aggressive compression lowers inference emissions and can justify applying KD in terms of carbon footprint, but may impose a quality ceiling, as observed for the 16M students. Future work should also examine synthetic-data filtering for Seq-KD. From a life cycle perspective such filtering is useful only if it reduces the amount of teacher generation required; filtering after the full synthetic corpus has already been generated may improve quality, but does not remove the main decoding footprint.
Overall, our {MT experiments} show that KD is not inherently greener: given the rapid developments of the AI industry amid an already tense climate situation, our research community has to put greater care into assessing how methods such as KD can actually help and when they fail to meet environmental constraints. 

\section*{Ethics and Impact Statement}
The environmental impact of machine learning models is complex to assess throughout the life cycle of software and hardware used to run it. It is a critical dimension of model evaluation that helps assess the feasibility and sustainability of models in addition to task performance. This study provides a comprehensive assessment of knowledge distillation for machine translation that contributes a comparative evaluation of distillation methods as well as a discussion of translation quality vs. environmental impacts trade-offs. 

This work is positioned within the purview of responsible science and sustainable computing by encouraging a broader scope of ML evaluation to drive development and deployment decisions. By providing guidelines for assessing the environmental impact trade-offs associated with different architectures, language pairs 
and level of use, this study fosters awareness and transparency within the MT and ML communities at large. 
While our empirical findings are specific to the MT settings studied here,  the life cycle accounting framework can be applied to other KD settings by re-estimating production and inference costs under the relevant functional unit and deployment regime. 

\section*{Limitations}

\paragraph{Hardware and deployment scope.}
Our measurements use a single NVIDIA V100 GPU and one data-center setting, with the grid emission factor, PUE, utilization assumptions, and hardware manufacturing footprint described in Section~\ref{sec:mlca} and Appendix~\ref{app:carbon_equations}. To address this, we report a sensitivity analysis for key accounting assumptions in Appendix~\ref{app:sensitivity}. We use a single GPU to keep comparisons controlled and avoid communication overhead; we report a preliminary four-GPU experiment in Appendix~\ref{app:distributed}.


\paragraph{Architecture coverage.}
The main MT experiments use TF-Big as the teacher and TF-Base/Tiny as students, allowing for a controlled comparison across KD methods. However, this does not cover all possible teacher--student architectures. Furthermore, we limit our study to transformers as prior work shows
they outperform recurrent models in KD for MT 
(e.g., higher BLEU scores~\cite{zhang-etal-2018-analyzing}).  Varying the architectures would introduce additional factors such as convergence behavior, capacity scaling, and optimization stability. However, our accounting framework is architecture-agnostic. In practice, most KD approaches in
MT have identical teacher and student architecture families~\cite{degibert2026kd4mtsurveyknowledgedistillation}, supporting our choice to avoid architectural variation.

\paragraph{Teacher-size selection.}
We use TF-Big as the main teacher. A larger 385M-parameter teacher did not improve COMET in our setting, consistent with the dependence of MT scaling benefits on data regime and possible quality saturation~\cite{gordon-etal-2021-data,zhang-etal-2023-towards-understanding,zhang-etal-2018-analyzing}. We therefore do not study teacher scaling systematically. A more systematic study of teacher scaling and KD efficiency is outside the scope of this study.

\paragraph{LLM comparability and teacher-cost allocation.}
The LLM experiment is preliminary and not directly comparable to the fully accounted encoder--decoder experiments. It uses Qwen3 models, LoRA adaptation, a 100K-sample distillation corpus, and sunk teacher-cost allocation because teacher pretraining emissions are unavailable, whereas the main MT experiments use teacher-included allocation. Real deployments may lie between these accounting extremes when teachers are reused across students, tasks, or language pairs, which would alter fixed production costs and break-even volumes. The LLM results should therefore be interpreted as a test of the same life cycle logic under more favorable assumptions rather than as a complete benchmark of LLM distillation for MT.

{\paragraph{Training-run variability.}
We train each KD configuration once. The paired-bootstrap confidence
intervals reported for COMET quantify uncertainty over the evaluation set,
but do not capture variation induced by training initialization or data
ordering. Consequently, our experiments do not establish the stability of
quality differences across training seeds. Repeating the full experimental
matrix across multiple seeds would provide stronger evidence on this
dimension and is left for future work.}

\paragraph{Evaluation coverage.}
Our experiments do not cover truly low-resource MT. Such settings may involve weaker teachers, greater sensitivity to data and optimization, less reliable automatic evaluation, and different production costs. In addition, we use COMET as the main translation-quality metric. Although COMET is widely used for MT evaluation, it does not capture all dimensions of translation quality, such as terminology consistency, robustness, hallucination, toxicity, or human preference.



\section*{Acknowledgments}
This work was supported in part by DataIA through a research visit by Joseph Attieh to LISN, by the GreenNLP project funded by the Research Council of Finland, by the Digital Europe Programme under grant agreement No. 101195233 (OpenEuroLLM), and by the French Agence Nationale de la Recherche (ANR) through grant InExtenso (ANR-23-IAS1-0004). The authors thank Clément Morand for his valuable insights on the use of MLCA in this study. The contents of this publication are the sole responsibility of its authors and do not necessarily reflect the opinion of the EU. The authors wish to acknowledge CSC – IT Center for Science, Finland, for computational resources. AI assistants were used for code debugging and editing.


\bibliography{custom}

\appendix
\section{Dataset Documentation}
\label{app:data_artifacts}

We use the constrained-track parallel corpora released for WMT 2024.  Figure~\ref{fig:corpus-sizes-all} reports the original parallel-corpus sources used for each language pair. We remove characters outside the scripts required for each language pair while preserving decimal digits, punctuation, and whitespace. For EN$\rightarrow$IS, we use the cleaned training corpus without down-sampling.  For EN$\rightarrow$HI and EN$\rightarrow$ES, we uniformly down-sample to 13M and 6M sentence pairs with seed 42 from the corresponding cleaned corpora. 

\begin{figure}[h]
  \centering
  \resizebox{\columnwidth}{!}{
  \begin{tikzpicture}
  \begin{groupplot}[
        group style={
            group size=1 by 3,
            vertical sep=1.5cm
        },
        xbar,
        width=0.5\textwidth,
        height=6cm,
        xmin=0,
        xlabel={Millions},
        xlabel style={font=\scriptsize},
        tick label style={font=\scriptsize},
        yticklabel style={font=\scriptsize, align=right},
        ytick=data,
        enlarge y limits=0.06,
        area legend,
    ]

\nextgroupplot[
        title={\scriptsize EN--ES},
        xmax=55,
        symbolic y coords={
            Wikipedia-v1.0,
            CommonCrawl-13,
            Europarl-7,
            Europarl-v8,
            DGT-v4,
            WikiMatrix-v1,
            DCEP-1,
            DGT-2019,
            WikiMatrix-FB,
            Wikititles-2014,
            UNPC,
            EuroPat-v3},
    ]
\addplot+[draw=cyan!70!black, fill=cyan!35]
coordinates {
    ( 1.81,Wikipedia-v1.0)
    ( 1.85,CommonCrawl-13)
    ( 1.96,Europarl-7)
    ( 2.01,Europarl-v8)
    ( 3.17,DGT-v4)
    ( 3.38,WikiMatrix-v1)
    ( 3.71,DCEP-1)
    ( 5.13,DGT-2019)
    ( 6.45,WikiMatrix-FB)
    (16.60,Wikititles-2014)
    (25.23,UNPC)
    (51.35,EuroPat-v3)
};

\nextgroupplot[
        title={\scriptsize EN--IS},
        xmax=9.5,
        symbolic y coords={
            {WikiMatrix-v1 (small)},
            WikiMatrix-1,
            ParIce EMA,
            EMA-2016,
            ELRC\_EMEA,
            XLENT-v1.1,
            CCAligned-v1,
            OpenSubtitles-2018,
            ParIce EEA,
            MultiParaCrawl-v7.1,
            ParaCrawl-9,
            CCMatrix-v1},
    ]
\addplot+[draw=orange!80!black, fill=orange!35]
coordinates {
    (0.09,{WikiMatrix-v1 (small)})
    (0.31,WikiMatrix-1)
    (0.40,ParIce EMA)
    (0.42,EMA-2016)
    (0.54,ELRC\_EMEA)
    (0.96,XLENT-v1.1)
    (1.19,CCAligned-v1)
    (1.57,OpenSubtitles-2018)
    (1.70,ParIce EEA)
    (2.39,MultiParaCrawl-v7.1)
    (2.97,ParaCrawl-9)
    (8.72,CCMatrix-v1)
};

\nextgroupplot[
        title={\scriptsize EN--HI},
        xmax=36,
        symbolic y coords={
            IE\_Sports,
            Anuvaad\_Judicial,
            OpenSubtitles-2018,
            Tanzil,
            Wikipedia-20210201,
            Business\_Standard,
            WikiMatrix-1,
            IITB-v2,
            IITB-train-1.5,
            XLENT-v1.1,
            MultiCCAligned-v1,
            AllenAI-NLLB-1},
    ]
\addplot+[draw=green!60!black, fill=green!35]
coordinates {
    ( 0.07,IE\_Sports)
    ( 0.08,Anuvaad\_Judicial)
    ( 0.09,OpenSubtitles-2018)
    ( 0.19,Tanzil)
    ( 0.19,Wikipedia-20210201)
    ( 0.27,Business\_Standard)
    ( 0.70,WikiMatrix-1)
    ( 1.56,IITB-v2)
    ( 1.56,IITB-train-1.5)
    ( 2.03,XLENT-v1.1)
    ( 8.18,MultiCCAligned-v1)
    (33.19,AllenAI-NLLB-1)
};

  \end{groupplot}
  \end{tikzpicture}}
  \caption{Size, in millions of sentence pairs, of the parallel-corpus sources for each language pair.}
  \label{fig:corpus-sizes-all}
\end{figure} 

{ After observing the substantial cost of training the EN$\rightarrow$IS setup, these sizes were chosen to create three distinct corpus scales while keeping the full set of teacher-training and distillation experiments feasible on a single GPU. Using the complete EN$\rightarrow$HI and EN$\rightarrow$ES corpora would substantially increase both teacher training and distillation costs. We additionally verified that the resulting teachers remain competitive with similarly sized OPUS-MT systems~\citep{tiedemann-thottingal-2020-opus}.
}

\section{Knowledge Distillation Objectives}
\label{app:kd_objectives}

\paragraph{Notations.} We consider autoregressive MT: given a source sentence $\mathbf{x}$, a model produces a target sequence
$\mathbf{y}=(y_1,\dots,y_T)$ by predicting each token $y_{i}$ conditioned on the source $x$ and the prefix $\mathbf{y}_{<i}$.
Let $D=\{(\mathbf{x}^{(i)},{\mathbf{y}}^{(i)})\}_{i=1}^{N}$ be a parallel corpus, where $x^i$ is a sentence in the source language and $y^i$ is the translation in the target language.  Knowledge distillation (KD) transfers knowledge from a
larger teacher to a smaller student~\cite{bucila2006model,hinton2015distilling}; in MT, the two
standard KD families are word-level (distribution matching) and sequence-level (synthetic targets)~\cite{kim-rush-2016-sequence}. 

We denote by $\theta_{\mathfrak{T}}$ and $\theta_{\mathfrak{S}}$ the parameter vectors of teacher and student. For a model of parameters $\theta$, the model's next-token distribution at time step $t$ for a specific example $(\mathbf{x},\mathbf{y})$ can be expressed as:
$p_{\theta}
\;:=\;
p(y_t \mid \mathbf{x}, \mathbf{y}_{<t}; \theta)$
.

\paragraph{Sequence-level KD (Seq-KD).}
Seq-KD~\cite{kim-rush-2016-sequence} trains the student on teacher-generated translations. For each source sentence $\mathbf{x}$, we decode a
single high-probability teacher output $\hat{\mathbf{y}}$ and form a synthetic corpus
$\hat{D}=\{(\mathbf{x}^{(i)},\hat{\mathbf{y}}^{(i)})\}_{i=1}^{N}$.

The synthetic target approximates the teacher argmax,
$\hat{\mathbf{y}} \approx \arg\max_{\mathbf{y}} p_{\theta_{\mathfrak{T}}}(\mathbf{y}\mid \mathbf{x})$,
typically computed with beam search. At each time step $t$, beam search maintains a set 
of $B$ partial hypotheses, expands them using the teacher
next-token distribution, and keeps the top-$B$
candidates by accumulated log-probability (optionally with a length penalty); the final output $\hat{\mathbf{y}}$ is
the highest-scoring completed hypothesis. Since decoding cost scales 
with $B$
, this stage can dominate the distillation compute.

Training is done through standard cross-entropy with teacher forcing:
\begin{equation*}
\mathcal{L}_{\mathrm{CE}}(\theta_{\mathfrak{S}};\hat{D})
\;=\;
\sum_{(\mathbf{x},\hat{\mathbf{y}})\in \hat{D}}
\sum_{t=1}^{|\hat{\mathbf{y}}|}
\mathcal{H}\!
(
\delta_{\hat{y}_t},
p_{\theta_{\mathfrak{S}}} 
)
\label{eq:seq_kd_ce}
\end{equation*}
Seq-KD is therefore an offline hard-distillation regime: the teacher supplies one-hot synthetic targets, and
additional compute is primarily incurred during teacher-side decoding.

\paragraph{Word-level KD (Word-KD).}
Word-KD~\cite{kim-rush-2016-sequence} trains the student to match the teacher's token-level distribution at each time step.

The dataset-level KD objective can be described as:
\begin{equation*}
\mathcal{L}_{\mathrm{KD}}(\theta_{\mathfrak{S}};\theta_{\mathfrak{T}})
\;=\;
\sum_{(\mathbf{x},\mathbf{y})\in D}
\sum_{t=1}^{|\mathbf{y}|} \KL(p_{\theta_{\mathfrak{T}}} \;\|\;p_{\theta_{\mathfrak{S}}})
\end{equation*}
In practice, student training combines teacher supervision with standard likelihood training on ground-truth targets
using an interpolation hyperparameter $\alpha\in[0,1]$:
\begin{equation*}
\resizebox{.95\linewidth}{!}{$
\mathcal{L}_{\textsc{Word-KD}}(\theta_{\mathfrak{S}};\theta_{\mathfrak{T}})
\;=\;
(1-\alpha)\,\mathcal{L}_{\mathrm{CE}}(\theta_{\mathfrak{S}})
\;+\;
\alpha\,\mathcal{L}_{\mathrm{KD}}(\theta_{\mathfrak{S}};\theta_{\mathfrak{T}})
$}
\label{eq:word-kd}
\end{equation*}
In the online form, teacher distributions are computed during student training, requiring repeated teacher forward passes. In offline variants, teacher distributions are precomputed and stored, often as top-$K$ logits or probabilities. This shifts part of the cost to a preprocessing stage but still requires running the teacher over the distillation corpus.

\section{Model, Tokenization, and Training Details}
\label{app:training_details}

\paragraph{Shared setup.}

Table~\ref{tab:shared_training_settings} summarizes the settings shared by the main encoder--decoder MT experiments. All KD methods are implemented in Fairseq, which allows us to reuse a single teacher checkpoint across distilled students and maintain consistent decoding, training, and evaluation pipelines across KD variants. We train one shared SentencePiece tokenizer~\cite{kudo-richardson-2018-sentencepiece} per language pair with a 32K BPE vocabulary and reuse it for teacher, No-KD, and KD runs. 
This avoids confounding KD comparisons with segmentation differences and is required for word-level KD, where teacher and student token distributions must be defined over the same vocabulary.  We enable byte fallback to improve coverage of rare characters and out-of-vocabulary symbols.

\begin{table}[h]
\centering
\small
\resizebox{\columnwidth}{!}{
\begin{tabular}{ll}
\toprule
\textbf{Component} & \textbf{Setting} \\
\midrule
Framework & Fairseq~\cite{ott-etal-2019-fairseq} \\
Optimizer & Adam~\cite{kingma2015adam} \\
Learning-rate schedule & Inverse square root \\
Tokenizer & SentencePiece BPE  \\
Vocabulary size & 32K shared source--target vocabulary \\
Validation set & FLORES+ dev \\
Test set & FLORES+ devtest \\
Quality metric & COMET, \texttt{Unbabel/wmt22-comet-da} \\
Early stopping & Based on FLORES+ dev performance \\
Hardware & Single NVIDIA V100 GPU \\
\bottomrule
\end{tabular}}
\caption{Shared experimental settings across the main encoder--decoder MT experiments.}
\label{tab:shared_training_settings}
\end{table}

\paragraph{Model architectures.}
Table~\ref{tab:arch} reports the teacher and student configurations for the main MT experiments. We use a Transformer-Big teacher and two student capacities: Transformer-Base for moderate compression and Transformer-Tiny for aggressive compression. 
The 65M student retains 31.1\% of the teacher parameters, while the 16M student retains 7.65\%. 
This design lets us compare KD behavior under both quality-preserving and capacity-constrained compression regimes.
\begin{table}[h]
  \centering
  \resizebox{\columnwidth}{!}{
\begin{tabular}{@{}lrrr@{}}
\toprule
 & \multicolumn{1}{c}{\textbf{Teacher}} & \multicolumn{2}{c}{\textbf{Students}} \\
\cmidrule(lr){3-4}
\textbf{Model} & \textbf{TF Big} & \textbf{TF Base} & \textbf{TF Tiny} \\
\midrule
Encoder layers        & 6     & 6     & 6 \\
Decoder layers        & 6     & 6     & 2 \\
Embedding dim         & 1024  & 512   & 256 \\
FFN dim               & 4096  & 2048  & 1536 \\
Attention heads       & 16    & 8     & 8 \\
\midrule
Parameters (M)        & 209   & 65    & 16 \\
\midrule
Params \% (w.r.t.\ Teacher) & -- & \textbf{31.1\%} & \textbf{7.65\%} \\
Compression Ratio & -- & \textbf{3.21} & \textbf{13.06} \\
\bottomrule
\end{tabular}}
\caption{Teacher and student configurations and compression ratios.}
 \label{tab:arch}
\end{table}

\paragraph{KD implementations and code release.}
Table~\ref{tab:method_code_sources} reports the source code for each method. For methods without a maintained public implementation matching our Fairseq pipeline, we provide our own.

\begin{table}[h]
\centering

\small\resizebox{\columnwidth}{!}{
\begin{tabular}{ll}
\toprule
\textbf{Method } & \textbf{Implementation}  \\
\midrule
Teachers / No-KD   & \href{https://github.com/facebookresearch/fairseq}{GitHub} \\
Seq-KD~\cite{kim-rush-2016-sequence}  & \href{https://github.com/facebookresearch/fairseq}{GitHub} \\
Seq-Inter~\cite{kim-rush-2016-sequence} & \href{https://github.com/Helsinki-NLP/CarbonFootprint.git}{GitHub*}   \\
Seq-Rep~\cite{zhang-etal-2018-analyzing}  & \href{https://github.com/Helsinki-NLP/CarbonFootprint.git}{GitHub*}  \\
Word-KD~\cite{kim-rush-2016-sequence}  &  \href{https://github.com/songmzhang/NMT-KD.git}{GitHub}\\
Sel-KD~\cite{wang-etal-2021-selective}  & \href{https://github.com/LeslieOverfitting/selective_distillation.git} {GitHub}\\ 
TIE-KD~\cite{zhang-etal-2023-towards-understanding}  & \href{https://github.com/songmzhang/NMT-KD.git}{GitHub} \\
Carbon accounting  & \href{https://github.com/Helsinki-NLP/CarbonFootprint.git}{GitHub*}   \\
\bottomrule
\end{tabular}}
\caption{Implementation sources and release plan. An asterisk (*) indicates implementations developed by the authors based on the corresponding published methods.}
\label{tab:method_code_sources}
\end{table}

\section{Carbon Accounting Details}
\label{app:carbon_details}

\subsection{Accounting Equations}
\label{app:carbon_equations}

MLCA evaluates the environmental impact of the system within the defined boundary, aiming to address the FU. As mentioned previously, we estimate the carbon footprint over the life cycle of the machine translation system. 
For each compute phase $\phi \in \{\textsc{train},\textsc{distill},\textsc{infer}\}$ and device $e$, we estimate the total  environmental impact for each system as the sum of operational and embodied emissions:  $I \;=\; \sum_{\phi}\sum_{e} \Bigl( I^{(\phi)}_{\mathrm{use},e} \;+\; I^{(\phi)}_{\mathrm{emb},e} \Bigr)$. 

\paragraph{Operational Emissions.}
The environmental impact of operational emissions $I^{(\phi)}_{\mathrm{use, e}}$ (CO$_2$e) corresponds to the greenhouse-gas emissions associated with the electricity used to run the workload of the phase $\phi$ on device $e$~\cite{berthelot2024}. It is estimated by first measuring the workload's electricity consumption of running phase $E^{(\phi)}$(kWh) on device $e$ \footnote{
When direct energy measurements are unavailable, $E^{(\phi)}$ is approximated from the average GPU power draw $P^{(\phi)}$ (kW) and wall-clock runtime $t^{(\phi)}$ (hours): $E^{(\phi)} \approx P^{(\phi)} \times t^{(\phi)}$~\cite{lacoste2019quantifyingcarbonemissionsmachine,henderson2020systematic}.}. 
Since the data-center hosting the device also spends energy on cooling and infrastructure, this energy is scaled with the  Power Usage Effectiveness (PUE) of the data-center~\cite{BRADY2013155}. Finally, this effective energy is converted to emissions by multiplying by the grid emission factor at the data-center location $g$, denoted $\mathrm{EGM}_g$ (kgCO$_2$e/kWh), reflecting the carbon footprint of the grid the device is plugged into \footnote{Because the electricity mix used by a specific data center is rarely disclosed, the workload is assumed to draw from the local grid of the data center's physical location, and publicly available regional emission factors are used to approximate $\mathrm{EGM}_g$~\cite{lacoste2019quantifyingcarbonemissionsmachine}.}. 
\paragraph{Embodied Emissions}
The environmental impact of the embodied emissions for device $e$, denoted $I^{(\phi)}_{\mathrm{emb},e}$, is estimated by amortizing the manufacturing footprint of the device $F^{(\mathrm{prod})}_e$ (kgCO$_2$e) over its expected lifetime ($t_{\mathrm{life},e}$) proportionally to the time the device is used in phase $\phi$ ($t^{(\phi)}_e$), and adjusted by its active utilization rate $\mathrm{AUR}_e \in (0,1]$~\cite{luccioni2024power,berthelot2024}.

\subsection{Parameter Values}
\label{app:carbon_values}

We use parameters consistent with prior MLCA/LCA work~\cite{luccioni2023,berthelot2024}. 
We assume an active utilization rate of 0.8 during training and distillation, and 0.2 during evaluation and inference, shown in Table~\ref{tab:carbon_parameters}. 
We assume a hardware lifetime of 5 years. 
The grid emission factor is set to the value corresponding to the data-center location (in our case, we used Puhti from CSC, which is located in Finland), $\mathrm{EGM}_g=0.033$ kgCO$_2$e/kWh. 
The PUE is set to the value reported by the data center, 1.24. 
For embodied emissions, we assume a manufacturing footprint of 150 kgCO$_2$e per GPU, consistent with \citet{falk2025carboncradletograveenvironmentalimpacts}.

\begin{table}[h]
\centering
\small
\resizebox{\columnwidth}{!}{
\begin{tabular}{@{}p{0.54\columnwidth}p{0.3\columnwidth}p{0.2\columnwidth}@{}}
\toprule
\textbf{Parameter} & \textbf{Value} & \textbf{Emissions} \\
\midrule
GPU type & NVIDIA V100 & All  \\
Hardware lifetime $t_{\mathrm{life},e}$ & 5 years & Embodied \\
AUR, training/distillation & 0.8 & Embodied \\
AUR, evaluation/inference & 0.2 & Embodied \\
Manufacturing footprint $F^{(\mathrm{prod})}_e$ & 150 kgCO$_2$e/GPU & Embodied \\
PUE & 1.24 & Operational\\
Grid emission factor $\mathrm{EGM}_g$ & 0.033 kgCO$_2$e/kWh & Operational\\
\bottomrule
\end{tabular}}
\caption{Carbon-accounting parameters used in the MLCA computation.}
\label{tab:carbon_parameters}
\end{table}

\subsection{Sensitivity Analysis}
\label{app:sensitivity}
\begin{figure}[h]
         \vspace{6mm}
         \includegraphics[width=0.48\textwidth]{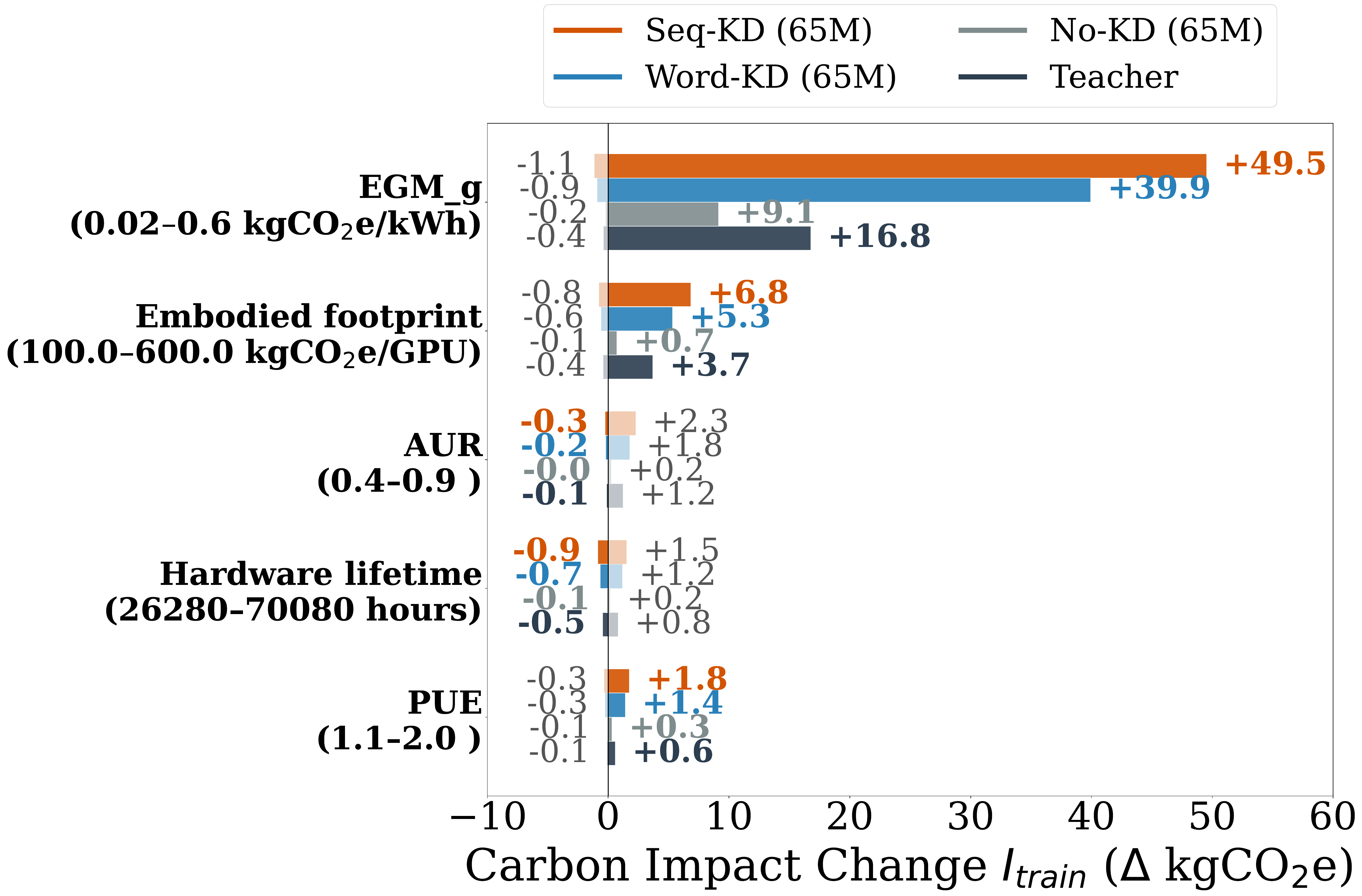}
     \caption{One-way sensitivity analysis of carbon impact for different setups. Bars represent the deviation ($\Delta$ kgCO$_2$e) from the best baseline when shifting parameters to the ranges indicated on the y-axis.}
     \label{fig:comparative_sensitivity}
\end{figure}

We report sensitivity analysis on the environmental impact of producing models, where the primary costs are one-time and MLCA assumptions directly affect relative footprints. Figure~\ref{fig:comparative_sensitivity} presents a one-way sensitivity analysis of the carbon footprint of producing different models under plausible ranges of key MLCA parameters.

 Grid carbon intensity is the dominant source of uncertainty, producing the largest absolute deviations across all methods (consistent with \citet{dodge2022}). In contrast, active utilization rate, hardware lifetime, and PUE induce relatively small changes to the carbon computation, as they affect only the amortized embodied component.  Importantly, the relative ordering of methods does not change across all parameter ranges: the No-KD model has the lowest production footprint, followed by the teacher, Word-KD, and Seq-KD. Larger models and distillation pipelines exhibit higher absolute sensitivity due to longer training times, but parameter uncertainty alters magnitudes rather than conclusions.

\section{Full Main MT Results}

\label{app:full_results}
\subsection{Total Footprint Decomposition}

\label{sec:rq1_cost}

\begin{figure}[ht]
         \centering
         \includegraphics[width=\columnwidth]{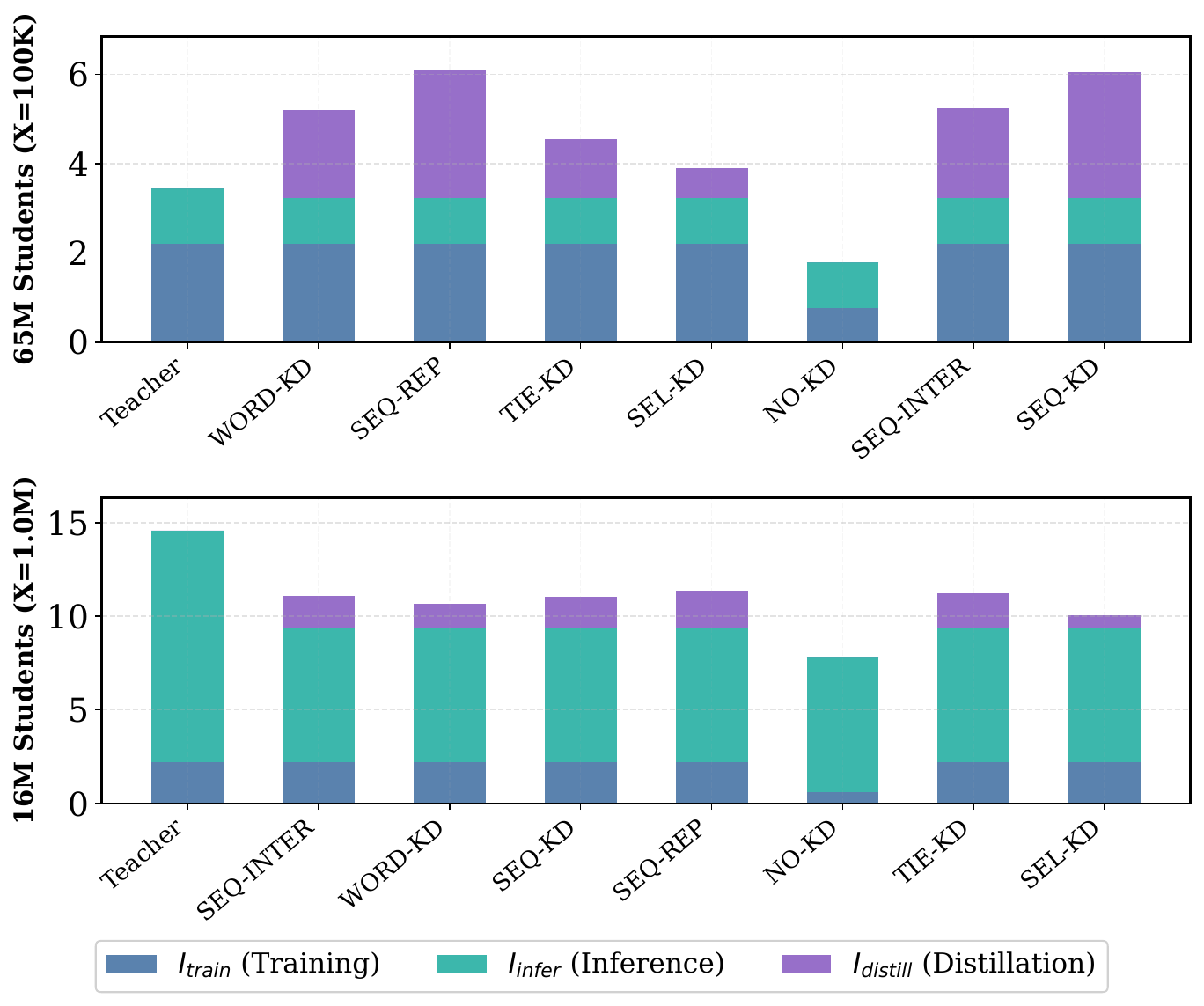}
     \caption{Total footprint (kgCO$_2$e) for all setups for a fixed served volume $X$ for $EN\rightarrow IS$. Non-teacher models are ordered by COMET score on FLORES+ devtest; scores are omitted for simplicity. }
     \label{fig:total_emissions}
\end{figure}

We quantify the total environmental impact 
for a fixed volume of $X$ tokens, as described in Section~\ref{sec:mlca}.  Figure~\ref{fig:total_emissions} reports the carbon footprint decomposed  into teacher training, distillation overhead, and inference emissions for EN$\rightarrow$IS. 

We present the results with two representative workloads: $X{=}100$K (roughly the size of a novel\footnote{For reference, \emph{Harry Potter and the Prisoner of Azkaban} has roughly 106,821 words, while \emph{Le Tour du monde en quatre-vingts jours} has around 70,000 words.}) and $X{=}1.0$M (corresponding to repeated usage, e.g., translating multiple novels). Since the No-KD baseline corresponds to training a model without KD, it does not involve teacher and distillation costs, unlike all other systems considered.

{Under this unbatched sentence-level serving configuration}, at $X{=}100$K, total emissions are largely determined by one-time costs. The emissions of the teacher and the No-KD baseline are the lowest as they both do not incur distillation emissions. For small translation workloads, KD is environmentally costly and difficult to justify over using the teacher. At $X{=}1.0$M, inference costs are the dominant costs for the teacher model. Using a compact student substantially reduces $I_{\text{infer}}(X)$, and KD can achieve lower total emissions than serving the teacher despite incurring the one-time distillation costs. This implies that 
KD can reduce total environmental impact once inference volume is high enough to amortize the fixed distillation cost.



\subsection{Break-Even Deployment Volumes}
\label{app:break_even}

Table~\ref{tab:break_even_points} reports the served token volume at which each distilled student has lower carbon emissions than the teacher. Break-even deployment volumes are reported in millions of decoded target tokens. 
An asterisk indicates methods that lie on the footprint--quality Pareto frontier. 
For non-frontier methods, the break-even point is computed using the checkpoint with the highest COMET score. 
Bold values indicate the Pareto-optimal method with the lowest break-even volume for each language pair and student size.
\begin{table}[h]
\centering
\small
\begin{threeparttable}
\begin{tabular}{lccc}
\toprule
\textbf{Method} & \textbf{EN$\rightarrow$ES} & \textbf{EN$\rightarrow$HI} & \textbf{EN$\rightarrow$IS} \\
& \textbf{M tokens} & \textbf{M tokens} & \textbf{M tokens} \\
\midrule
\multicolumn{4}{l}{\textbf{16M student}} \\
\midrule
Seq-Inter & 0.057$^\ast$ & \textbf{0.052}$^\ast$ & 0.595$^\ast$ \\
Seq-Rep   & \textbf{0.037}$^\ast$ & 0.104$^\ast$ & 0.382 \\
Sel-KD    & 0.469 & 0.461 & 0.130 \\
Seq-KD    & 0.152$^\ast$ & 0.181$^\ast$ & 0.407 \\
TIE-KD    & 0.952 & 0.791 & 0.350 \\
Word-KD   & 1.006 & 0.748 & \textbf{0.439}$^\ast$ \\
\midrule
\multicolumn{4}{l}{\textbf{65M student}} \\
\midrule
Seq-Inter & 0.199$^\ast$ & \textbf{0.091}$^\ast$ & 0.907 \\
Seq-Rep   & \textbf{0.160}$^\ast$ & 0.236$^\ast$ & 1.301 \\
Sel-KD    & 1.126$^\ast$ & 0.976 & 0.306 \\
Seq-KD    & 0.343$^\ast$ & 0.502$^\ast$ & 1.468 \\
TIE-KD    & 2.243 & 1.396 & \textbf{0.671}$^\ast$ \\
Word-KD   & 1.516 & 1.862 & 1.002$^\ast$ \\
\bottomrule
\end{tabular}

\caption{{Unbatched sentence-level} break-even deployment volumes for distilled students. Lower values indicate that the upfront distillation footprint is amortized after fewer translated tokens.}
\label{tab:break_even_points}
\end{threeparttable}
\end{table}

The table shows why break-even analysis must be interpreted together with quality. The 16M students usually break even earlier because they have lower inference footprints, but it comes with quality ceilings.  The 65M students require more served tokens to amortize production costs, but they have better translation quality. 
The lowest break-even method is therefore not automatically the best deployment choice; the selected checkpoint must also satisfy the required COMET level, and lie on or close to, the footprint--quality Pareto frontier.

\section{Distributed Distillation}
\label{app:distributed}

We evaluate whether distributing distillation across four GPUs changes the footprint--quality trade-off. In our setting, distributed distillation does not improve the Pareto frontier: the carbon impact of distillation increases by an average factor of approximately 3 for 65M students and 5 for 16M students. Runtime gains are insufficient to offset the additional operational and embodied footprint of using more devices. Therefore, the four-GPU setting is dominated by the single-GPU setup and is excluded from the main Pareto analysis.

\section{Decoder-Only LLM Case Study}
\label{app:llm_details}
This appendix reports implementation details for the decoder-only LLM case study. The COMET scores, carbon footprints, and break-even values are reported in the main text.

\subsection{Setup and Teacher-Cost Allocation}
We distill from Qwen3-8B and Qwen3-4B teachers into a Qwen3-0.6B student. 
All models are from the same Qwen3 family, which ensures a shared tokenizer and vocabulary. 
This is required for word-level/logit-based KD because teacher and student distributions must be defined over the same token vocabulary.

We evaluate EN$\rightarrow$IS and EN$\rightarrow$ES. For each direction, we sample 100K sentence pairs from the filtered WMT 2024 data used in the main experiments. This is five times the sample size used in CycleDistill~\cite{halder-etal-2025-cycledistill}. 
The same sampled subset is used for Seq-KD and Word-KD so that the methods differ only in the teacher signal provided to the student. 
Sampling is deterministic and approximately uniform over the filtered file: we count the total number of lines, set the interval as $N_{\mathrm{total}}/N_{\mathrm{sample}}$, and select examples at rounded interval positions while preserving the original corpus order.

Unlike the main encoder--decoder experiments, this case study uses sunk teacher-cost allocation, $\lambda_T=0$. 
This means that the pretraining footprint of the Qwen3 teacher is not assigned to the distilled student. 
We use this allocation because the full pretraining footprint of the open LLM teachers is unavailable and cannot be reconstructed reliably. 
The reported production footprint therefore includes teacher-side distillation preprocessing and student adaptation, but not teacher pretraining.

\subsection{Distillation and Training Details}
For Seq-KD, the teacher generates synthetic target translations for the sampled source sentences using greedy decoding. The student is then trained on the resulting synthetic parallel corpus.

For Word-KD, we use offline token-level distillation on only the top-$K$ teacher logits.  Following
 \citet{halder-etal-2025-cycledistill}, we store the top-$20$ teacher probabilities and token indices for each generated target position during teacher generation.   The probabilities are quantized to 8-bit values and saved together with the corresponding token indices.  The generated translations, references, source sentences, and file pointers are stored in a metadata file, while the top-$20$ arrays are stored separately as NumPy files. 
 
 This design avoids querying the teacher during every student-training step and avoids storing full-vocabulary logits, making Word-KD cheaper in both compute and storage than its online version. 
However, the savings come from replacing the full teacher distribution with a sparse top-$K$ approximation: the student receives high-probability alternatives, but not the complete long-tail distribution. 

The Qwen3-0.6B student is adapted with LoRA adapters. 
Table~\ref{tab:llm_lora_hparams} reports the shared LoRA hyperparameters used in the decoder-only LLM case study. 
We build on the CycleDistill training code~\cite{halder-etal-2025-cycledistill} and modify it to support FLORES-based validation and early stopping. 

\begin{table}[h]
\centering
\small
\resizebox{\columnwidth}{!}{
\begin{tabular}{ll}
\toprule
\textbf{Hyperparameter} & \textbf{Value} \\
\midrule
Student model & Qwen3-0.6B \\
Adaptation method & LoRA \\
Target modules & 
\makecell[l]{\texttt{q\_proj}, \texttt{k\_proj}, \texttt{v\_proj}, \texttt{o\_proj},\\
\texttt{gate\_proj}, \texttt{up\_proj}, \texttt{down\_proj}} \\
LoRA rank $r$ & 32 \\
LoRA scaling $\alpha_{\mathrm{LoRA}}$ & 64 \\
LoRA dropout & 0.05 \\
Learning rate & $10^{-4}$ \\
Per-device batch size & 2 \\
Gradient accumulation & 16 \\
Maximum sequence length & 2048 \\
\bottomrule
\end{tabular}}
\caption{LoRA adaptation hyperparameters for the decoder-only LLM case study.}
\label{tab:llm_lora_hparams}
\end{table}

For decoder-only LLM generation, we use a fixed translation prompt for all language directions:
\begin{quote}
\small\ttfamily
Translate the following \{source\_language\} source text to \{target\_language\}.\\
Only output the translation. Do not explain.\\
\{source\_language\}: \{source\_text\}\\
\{target\_language\}:
\end{quote}
The placeholders \texttt{\{source\_language\}}, \texttt{\{target\_language\}}, and \texttt{\{source\_text\}} are replaced with the corresponding language names and source sentence. This prompt is used consistently across this case study.

\subsection{Comparability Caveat}
The LLM case study is intended as a scoped extension, not as a directly comparable benchmark against the main encoder--decoder MT experiments. 
It differs in four important ways: it uses decoder-only Qwen3 models rather than Transformer encoder--decoder MT models; it uses LoRA-based adaptation rather than full-model training; it uses a much smaller sampled distillation subset rather than multi-million-sentence corpora; and it uses sunk teacher-cost allocation because teacher pretraining emissions are unavailable. 
At the same time, this setting reflects a more practical deployment scenario: practitioners often start from an existing open LLM, adapt it with parameter-efficient fine-tuning, and treat the teacher pretraining cost as already sunk rather than retraining a teacher from scratch. 
For these reasons, the case study is used to test whether the same life cycle logic appears under a more realistic and favorable teacher-cost assumption.

\section{Artifact Licenses}
We use publicly available research artifacts under their respective licenses and terms of use, including the WMT 2024 constrained-track corpora, FLORES+, COMET, Fairseq code, SentencePiece, Qwen3 models, and CycleDistill code.

\end{document}